# PREDICTIVE MAINTENANCE IN PHOTOVOLTAIC PLANTS WITH A BIG DATA APPROACH

Alessandro Betti [†], Maria Luisa Lo Trovato [‡], Fabio Salvatore Leonardi [‡],
Giuseppe Leotta [‡], Fabrizio Ruffini [†] and Ciro Lanzetta [†]
[†] i-EM srl, via Aurelio Lampredi 45, Livorno (Italy)
[‡] Enel Green Power SPA, Viale Regina Margherita 125, Rome (Italy)

ABSTRACT: This paper presents a novel and flexible solution for fault prediction based on data collected from SCADA system. Fault prediction is offered at two different levels based on a data-driven approach: (a) generic fault/status prediction and (b) specific fault class prediction, implemented by means of two different machine learning based modules built on an unsupervised clustering algorithm and a Pattern Recognition Neural Network, respectively. Model has been assessed on a park of six photovoltaic (PV) plants up to 10 MW and on more than one hundred inverter modules of three different technology brands. The results indicate that the proposed method is effective in (a) predicting incipient generic faults up to 7 days in advance with sensitivity up to 95% and (b) anticipating damage of specific fault classes with times ranging from few hours up to 7 days. The model is easily deployable for on-line monitoring of anomalies on new PV plants and technologies, requiring only the availability of historical SCADA and fault data, fault taxonomy and inverter electrical datasheet.

## 1 INTRODUCTION

The provision of a Preventive Maintenance strategy is emerging nowadays as an essential field to keep high technical and economic performances of solar PV plants over time [1]. Analytical monitoring systems have been installed therefore worldwide to timely detect possible malfunctions through the assessment of PV system performances [2-3].

However, high customization costs, the need of collecting a great number of physical variables and of a stable Internet connection on field generally limit their effectiveness, especially for farms in remote places with unreliable communication infrastructures. The lack of a predictive component in the maintenance strategy is also a hindrance to minimize downtime costs. In order to keep lower the implementation costs and model complexity, statistical methods based on Data Mining are recently emerging as a very promising approach both for fault prediction and early detection. However, while the Literature is mainly focusing on equipment level failures in wind farms [4-5], research for PV plants is still in an early stage [6].

The present paper describes an innovative and versatile solution for inverter level fault prediction based on a data-driven approach, already tested with remarkable performances on six PV plants of variable size up to 10 MW located in Romania and Greece and three different inverter technologies (Table 1). The proposed approach is easily portable on different plants and technologies and simplifies the update process to follow the PV plant evolution.

## 2 METHODOLOGY

The model is composed by two parallel modules both being capable of predicting incipient faults, but differing for the level of details provided and the remaining operational time after first indication of fault: a Supervision-Diagnosis Model (SDM) based on a Self-Organizing Map (SOM) predicting generic failures through a measure of deviations from normal operation, and a Short-Term Fault Prediction Model (FPM) based on artificial Neural Network (NN) addressing the prediction of specific fault classes. The main steps of the model workflow are described in the following.

**Table 1:** list of tested PV plants.

| Plant Number | # of Inverter Modules | Inverter Manufacturer Number | Max Active Power (KW) | Plant Nominal Power (MW) |
|---|---|---|---|---|
| 1 | 35 | 1 | 385 | 9.8 |
| 2 | 7 | 1 | 385 | 2.8 |
| 3 | 2 | 2 | 731.6 | 1.63 |
| 4 | 25 | 3 | 183.4 | 4.9 |
| 5 | 34 | 3 | 183.5 | 6.0 |
| 6 | 10 | 3 | 183.4 | 1.99 |

### 2.1 Data and Alarm Logbook Import

Historical data extracted from 5-min averaged SCADA data (Table 2) and inverter manufacturer electrical parameters for the on-site inverter technology are first collected to train the model. SCADA logbooks, as well as fault taxonomy, are also imported for offline performance assessment and normal training period selection in the case of the SDM, and for NN training for the FPM. Logbook import is achieved by matching fault classes listed in the fault taxonomy, which includes also fault severity, to the fault occurrences recorded in logbooks and discretizing them on the timestamp grid of SCADA data. In particular, a $k$-th fault is assigned to timestamp $t_n$ if the following condition occurs:

$$t_{start,k} \leq t_n \leq t_{end,k}, \qquad (1)$$

where $t_{start,k}$ ($t_{end,k}$) are the initial (final) instant of fault event. Then SCADA data labelling is realized by assigning fault codes (as integer numbers) to SCADA data. In the case of concurrent fault events, a prioritization rule is adopted considering only the most severe fault and, if necessary, the most frequent fault in that day.

**Table 2:** electrical and environmental input SCADA data (GHI (GTI): Global Horizontal (Tilted) Irradiance).

| DC Electrical tags | AC Electrical tags | Temperature tags | Irradiance tags |
|---|---|---|---|
| Current ($I_{DC}$) | Current ($I_{AC}$) | Internal ($T_{int}$) | GTI |
| Voltage ($V_{DC}$) | Voltage ($V_{AC}$) | Panel ($T_{mod}$) | GHI |
| Power ($P_{DC}$) | Power ($P_{AC}$) | Ambient ($T_{amb}$) | |

2.2 Data Preprocessing

AC power ($P_{AC}$) depends primarily on the level of solar irradiance (GTI) and, secondarily, on ambient temperature ($T_{amb}$). A first-order regression of signals GTI and $P_{AC}$ applied on training set samples allows to remove outliers corresponding to the furthest points from fitting: $|P_{AC} - GTI \cdot m + b| > thr \cdot (GTI \cdot m + b)$, where $m$ and $b$ are the slope and the intercept, respectively, computed by a least squares approach and $thr$ is the threshold set by a trial and error process.

Further preprocessing steps include removal of days with a large amount of missing data, setting periodic tags to 0 in the nighttime, data range check and removal of unphysical plateaus.

2.3 Data Imputation

Missing instances in test set are imputed by means of a $k$-Nearest Neighbors ($k$-NN) algorithm using the training set as the reference historical dataset, selecting nearest neighbors according to the Euclidean distance and exploiting hyperbolic weights.

2.4 Feature Engineering: Data De-trending and Scaling

In order to remove the season-dependent variability from input data, a de-trending procedure has been applied by following customized approaches for each tag. $T_{mod}$ training data have been de-trended by means of the least squares method to find the best line $T_{fit}$ against $T_{amb}$ and selecting only low $GTI$ samples to remove the effect of panel heating due to sunlight:

$$T_{mod}^{detrend} = (T_{mod} - T_{fit})/T_{fit}, \quad (2)$$

where $T_{fit} = m_T \cdot T_{amb} + b_T$ is the fitting, $m_T$ is the regression slope and $b_T$ is the intercept. All the remaining input tags, but DC and AC voltages, have been de-trended according to a classical Moving Average (MA) smoothing to compute the trend component and applying an additive model for time series decomposition. In both cases also test instances have been de-trended by means of a moving window mask 1-day long for tracking of time-varying input patterns. Finally, input data normalization is performed to avoid unbalance between heterogeneous tags.

2.5 Data Processing: the Supervision Model (SDM)

SDM has been built by means of an unsupervised clustering approach based on a 20×20 Self-Organizing Map (SOM) algorithm (Figure 1) performing a non-linear mapping from a n-dimensional space (n=11) to a 2-dimensional space with an online weight update rule [7] and trained on a normal operation period, identified by a time interval, through a competitive learning process. SOM has the valuable property of preserving input topology: neighbor neurons in the SOM layer respond to similar input vectors. As a consequence, a change in the distribution of input instances, due for example to inverter malfunction, leads to a different data mapping in the output grid. A multivariate statistical process control analysis in the form of a control chart may be therefore built to detect this change in the patterns distribution.

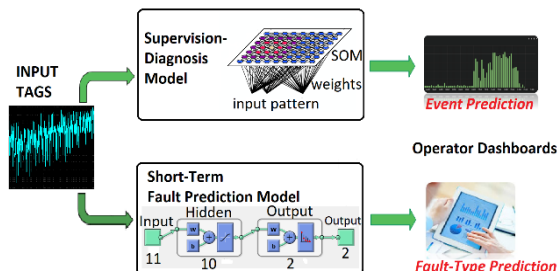

**Figure 1**: workflow of the Model SDM (SOM based) and FPM (NN based) from SCADA input tags to model outputs available in an operator dashboard.

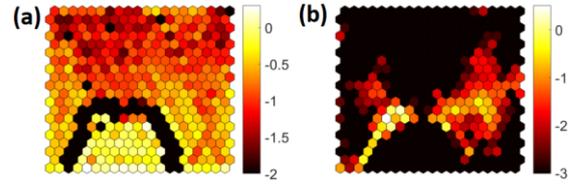

**Figure 2**: (a) $log(P_{ij}^{TRAIN})$ and (b) $log(P_{ij}^{TRAIN} - P_{ij}^{TEST-FULL})$ considering the full training and test phases. The larger the occupancy difference in (b), the smaller the contribution to KPI value (Eq. (3)).

A Key Performance Indicator (KPI) has been defined as in Eq. (3), measuring a process variation at generation unit level from the normal state towards abnormal operating conditions when a threshold is crossed:

$$KPI = \sum_{i,j} P_{ij}^{TEST} \cdot \frac{1 - |P_{ij}^{TRAIN} - P_{ij}^{TEST}|}{1 + |P_{ij}^{TRAIN} - P_{ij}^{TEST}|}, \quad (3)$$

where sums run on the SOM cells, $P_{ij}^{TRAIN} = N_{ij}^{TRAIN}/\sum_{ij} N_{ij}^{TRAIN}$ is the normalized number of input patterns mapped into the cell (i,j) in the training phase and $P_{ij}^{TEST} = N_{ij}^{TEST}/\sum_{ij} N_{ij}^{TEST}$ is the normalized number of input patterns mapped into the cell (i,j) in a 24-hours test phase (Figure 2).

The ratio factor in Eq. (3) penalizes large deviations from the training operating conditions: indeed if $\forall (i,j) \; P_{ij}^{TEST} = P_{ij}^{TRAIN}$ then $KPI = 1$, and if $|P_{ij}^{TRAIN} - P_{ij}^{TEST}| \to 1$ ($P_{ij}^{TRAIN} = 0, P_{ij}^{TEST} = 1$, or viceversa), then $KPI \to 0$. From a physical point of view, Eq. (3) is a robust indicator detecting changes in the underlying non-linear dynamics of the generation unit from normal status, represented by $KPI = 1$.

To remove the seasonality pattern from Eq. (3) due to time-varying number of daytime instances, a KPI de-trending procedure has been followed by means of a linear regression of both signals to compute the KPI trend and selecting the de-trended component $KPI/KPI_{trend}$ by assuming a multiplicative decomposition $KPI = KPI_{trend} \times KPI_{season} \times KPI_{cyclic} \times KPI_{residual}$. Once the de-trended KPI is available, a low-pass MA filter with a 4-weeks long window is applied. Finally, four warning levels of different severity are computed based on thresholds crossing and time persistence rules (Table 3). Model performances have been evaluated by means of usual classification metrics (accuracy, sensitivity and specificity) exploiting the alarm logbook knowledge and assigning a predictive connotation to sensitivity by assuming correctly predicted a fault event if in the last $N$ days prior to fault a warning has been triggered by the SDM.

**Table 3:** rules for switching ON of the 4 warning levels (w) of the SDM (d: day).

| Warning Level (w) | Crossing of Threshold | KPI Derivative | Persistence |
|---|---|---|---|
| 1 | $3\sigma_{KPI^{TRAIN}}$ | < 0 | 1 d |
| 2 | $3\sigma_{KPI^{TRAIN}}$ | < 0 | 2 d of w 1 |
| 3 | $5\sigma_{KPI^{TRAIN}}$ | < 0 | 1 day |
| 4 | $5\sigma_{KPI^{TRAIN}}$ | < 0 | 2 d of w 3 |

2.6 Data Processing: the Short-Term Fault Prediction Model (FPM)

Short-Term FPM is based on a 11-10-2 Pattern Recognition Feed-Forward Back-Propagation NN fed with 11 input tags (Table 2) and containing one hidden layer with 10 neurons and two output neurons (Figure 1) trained to recognize specific fault classes by means of a Bayesian regularization algorithm to prevent overfitting. NN architecture has been optimized building up an ensemble of statistical simulations to maximize classification metrics. Once the full dataset is labelled (Section 2.1) and preprocessed (Section 2.2), a data sampling step is followed, to compensate the large unbalance between the number of normal operation data of majority class and the one of low frequency failure data, causing a prediction bias affecting event classification.

Sample balancing is achieved by first collecting the number of fault instances $N_{fault,i}$ available for the $i$-th fault class and then assigning $2/3$ of them to training set, which is finally filled by randomly sampling normal instances up to the percentage required for almost balanced classes. Test set is instead built by means of the remaining $1/3 \times N_{fault,i}$ fault instances and randomly sampling normal instances up to the proportion necessary for representing the distribution of the labeled SCADA data. Normal and fault samples are sampled separately to avoid unbalanced fault distribution in training and test set.

Specifically, for prediction from current time $t$ an event occurs (either faulty or normal) to the previous $n$ hours a different training (test) set has been built by considering instances at time $t$ $(t-n)$. In this manner, NN is trained to recognize events at time they occur ($t$) and tested at previous instants ($t-n$). Due to the poor fault statistics available, missing data at time $t-n$ have been imputed by means of a $k$-NN algorithm. Model performances have been finally assessed by setting-up a Monte Carlo approach for each timestamp and averaging ensemble classification metrics.

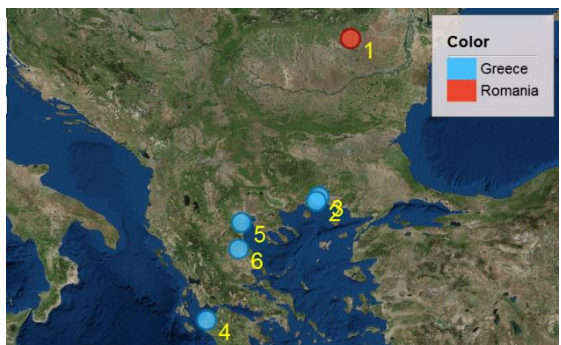

**Figure 3**: position of PV plants. Plants #2 and #3 in Greece are close to each other and their corresponding circles overlap in the figure.

## 3 RESULTS AND DISCUSSION

SDM has been tested on six different PV plants located in Romania and Greece (Figure 3) and corresponding, globally, to more than one hundred inverters of three well-known technology brands and typologies (inverter module, central inverter, master slave). The time period considered for offline performance assessment spans from 2014 to 2016, depending on data source availability.

In Figure 4 the KPI, warning levels, and fault occurrences percentage time series are shown for inverter A.3 of plant #1 in Romania with installed capacity of almost 9.8 MW and inverter technology #1. The two thresholds $3\sigma_{KPI^{TRAIN}}$ and $5\sigma_{KPI^{TRAIN}}$ are also represented by dashed and dotted black curves, respectively. According to alarm logbooks available, a series of thermal issue damages happened on different inverters in 2014-2015 which led to inverter replacements. In particular, a generalized failures occurred from 3rd to 28th of November 2014 to inverter A (AC Switch Open, severity 2/5) but it was recognized only later by the operator, with a severe lost production. As can be seen,

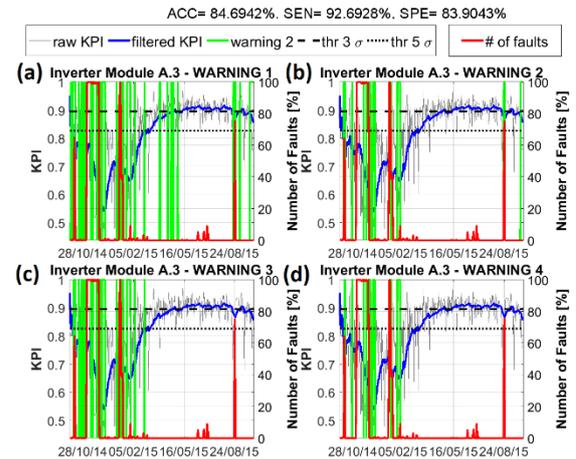

**Figure 4**: raw (sky-blue) and filtered (blue) KPIs, warning levels (green) and fault occurrences percentage (red, on right *side*) as a function of time for INV. A.3 of PV plant #1 (9.8 MW) and technology #1. Warnings 1 (a) to 4 (b) are shown from top left to bottom right for period October 2014 to August 2015.

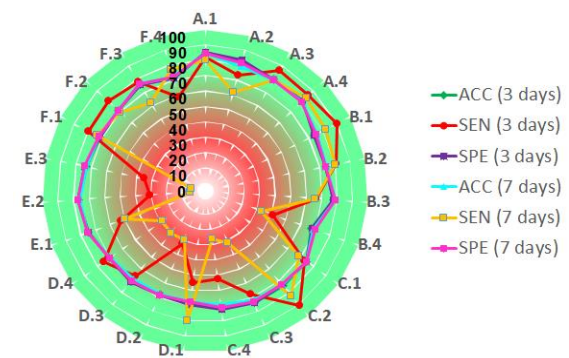

**Figure 5**: radar chart of classification metrics, expressed as a percentage 0-100% (ACC: accuracy, SEN: sensitivity, SPE: specificity), computed at 3 and 7 days prior of fault occurrences for inverter A-F of PV plant #1. Inverters G-H are neglected since no faults happened.

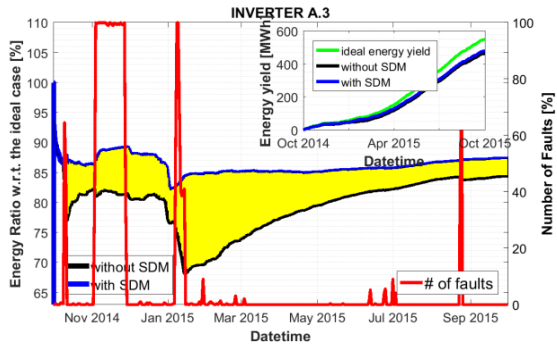

**Figure 6**: energy yield time series w.r.t. ideal case with (blue curve) or without (black) the predictive service: yellow area represents the energy gain enabling SDM. Fault occurrences percentage is also shown on the right (red). Inset: energy yield in the ideal case, as well as with or without SDM (INV. A.3 of Plant #1).

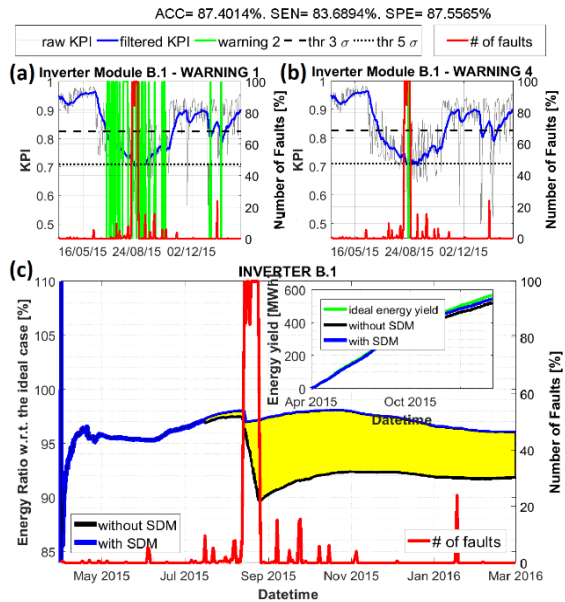

**Figure 7**: raw and filtered KPIs, as well as normalized fault occurrences and warning levels 1 (a) and 4 (b) for INV. B.1 of PV plant #2 (2.8 MW) and technology #1. (c): energy yield time series with respect to the ideal case with or without SDM enabled, as well as fault occurrences percentage as a function of time. Inset: energy yield in the ideal case, and with or without SDM.

Supervision Model well anticipates logbook fault events, with a clear correlation between the deep KPI degradation (with warning triggered up to level 4) and fault occurrences, even for events happening in 7-10 January 2015 (DC Ground Fault, i.e. high leakage current to ground on DC side, severity 1/5) and 23-24 August 2015 (DC Insulation Fault, i.e. overvoltage across DC capacitors, severity 2/5). Sensitivity roughly degrades from 93% to 84% anticipating faults from 0 up to 7 days, and with an overall accuracy of almost 85%.

Figure 5 shows a general overview of SDM performance on plant #1 at 3 and 7 days in advance with respect to fault occurrences. As can be seen, at 3 (7) days sensitivity (SEN) is larger than 60% for 17 (14) devices (on a total of 23), with a mean sensitivity of 72% (61%) at 3 (7) days. Accuracy (ACC) and specificity (SPE) are instead, on average, almost 80% for both the considered time horizons.

An estimate of the production gain achievable by means of a predictive service may be obtained by computing lost production as:

$$LP(t) = \int_o^t [P_{th}(t') - P_{AC}(t')] \, dt' \,, \quad (4)$$

where $P_{th} = P_{nom}[1 - k_{pv}] \cdot GTI/GTI_{SC}$ is the theoretical power in normal condition [8], $GTI_{SC} = 1200 \, W/m^2$ is the irradiance in standard conditions and $k_{pv} = (T_{cell} - 25)\frac{\gamma}{100}$ if $T_{cell} \geq 25°C$ if $T_{cell} \geq 25°C$ and 0 otherwise. In Figure 6 the energy yield computed with respect to the ideal case ($P_{AC}(t) = P_{th}(t) \, \forall t$) is presented as a function of time for the case with (assuming $P_{AC} = P_{th}$ if a fault is correctly predicted) and without SDM enabled. As can be seen, an energy yield improvement (yellow area) up to 10-15% may be achieved with SDM.

In Figure 7a-b the KPI and warning levels 1 and 4, respectively, are illustrated as a function of time for inverter B.1 of PV plant #2 (2.8 MW) with installed technology #1. As may be observed, the strong fault spike occurred in August 2015 (red curve) due to an input over-current on DC side (severity 2/5) is well predicted by SDM, with a sensitivity larger than 95% and an overall accuracy above 80% event at 7 days in advance. The energy yield gain achievable when applying SDM is almost 6-7% (Figure 7c).

Some fault classes cannot be predicted due to their instantaneous nature. SDM, thanks to its parametric structure with respect to inverter technology and plant configuration and to an ad-hoc tuning of model parameters on each specific plant, guarantees early detection for these faults. In Figure 8 the case of plant #4 in Greece (4.9 MW) with installed inverter technology #3 is reported. As can be seen, in the second half of May 2016, SDM early detects a severe anomaly at inverter 3.5 due to an IGBT stack fault which led to inverter replacement. Due to the sharp KPI decrease below threshold $5\sigma_{KPI^{TRAIN}}$, warnings 1 and 3 are triggered almost instantaneously, whereas warnings 2 and 4 with a delay of almost 2 days.

Short-Term FPM model has been applied to three PV plants (#1, #2, #3) and two different inverter technologies (#1 for plants #1, #2 and #2 for plant #3). In Figure 9 the classification metrics, as well as the number of detected faults, are shown as a function of the time prior to the fault occurrence for the highest frequency failure class of PV plants #1, #2 and #3 at fixed inverter module. As can be seen in Figure 9a, when thousands of occurrences are available, an outstanding prediction capability is achieved with sensitivity decreasing down to value of almost 50-60% seven days in advance. According to State of the Art [4], prediction performances degrade generally much faster on time horizon ranging from 1 hour to 12 hours for low frequency failures with less than 100 fault instances available (Figure 9b), with some exceptions depending on correlations among predictors and faults (Figure 9c).

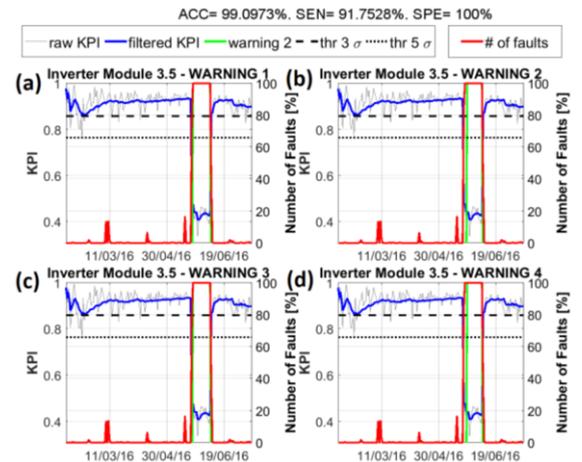

**Figure 8**: raw and filtered KPIs, as well as normalized fault occurrences and warning levels 1 (a) – 4 (d) for INV. 3.5 of PV plant #4 (4.9 MW) and technology #3.

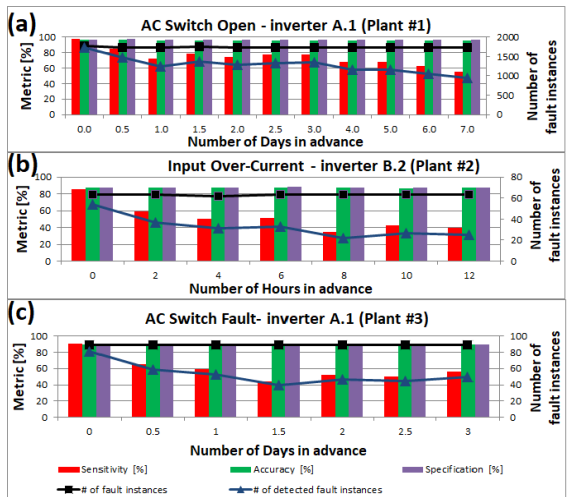

**Figure 9**: classification metrics (bar plot on the left) and number of faults and of detected faults (black and blue curves on the right, respectively) as a function of time in advance. (a): fault class AC Switch Open (plant #1); (b): fault class Input Over-Current (plant #2); (c): fault class AC Switch Fault (plant #3).

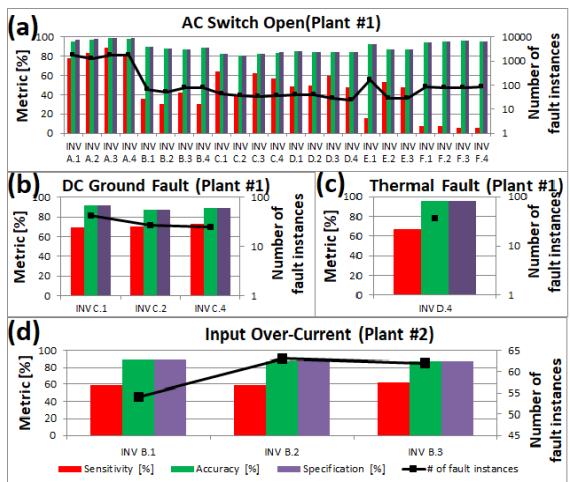

**Figure 10**: classification metrics (bar plot on the left) and number of fault instances (black curve on the right) as a function of inverter module for fault classes (a) AC Switch Open, (b) DC Ground Fault, (c) Thermal Fault (Low $T_{amb}$) of plant #1 computed at t-3Days, and (d) for class Input Over-Current of plant #2 calculated at t-2Hours.

In Figure 10 a global FPM performance overview on the inverter park is shown for classes AC Switch Open, DC Ground Fault and Thermal Fault (Low Ambient Temperature) of PV plant #1 computed 3 days prior of the failure, and for class Input Over-Current for PV plant #2 calculated 2 hours before the damage. In general, FPM is capable of detecting incipient faults for up to three or four fault classes for each inverter module. Figure 10a highlights however a strong correlation between FPM performances and failure data availability, with sensitivity (red bars) degrading dramatically from 80% (inverter A) to roughly 30-40% (inverter B-F), on average, when the number of fault instances (black curve) decreases from thousands to one hundred or less.

Accuracy (green bars) is still over 80% due to good negative samples classification performances. Figure 10b-d confirm previous conclusions: when few fault instances (roughly 30-40) are available, sensitivity is satisfactory generally on very short time horizons (2 hours in Figure 10d), but in some cases even larger (3 days in Figure 10b-c) depending on the strength of correlations among predictors and failure data.

## 4 FINAL REMARKS

An original methodology to predict inverter faults at two different levels of detail (prediction of a status/fault, prediction of a specific fault) has been presented and validated on SCADA data collected from 2014 to 2016 on a park of six PV plants up to 10 MW.

Results demonstrates that the proposed SDM effectively anticipates high frequency inverter failures up to almost 7 days in advance, with sensitivity up to 95% and specificity of almost 80%. SDM also guarantees early detection for unpredictable failures. FPM exhibits also excellent predictive capability for high frequency fault classes up to 7 days prior of the damage but, depending on fault statistics available, sensitivity may degrades also on horizons of few hours. Combination of these two prediction modules can therefore enable PV system operators to move from a traditional reactive maintenance activity towards a proactive maintenance strategy, improving decision-making process thanks to a complete information on the incoming failure before the fault occurs. The model is actually tested for on-line monitoring of anomalies in Romania and Greece and can be easily deployed on new PV plants thanks to the limited amount of information required.

Next steps may include the introduction of a deep-learning based automated system of fault detection in drone-based thermal images of PV modules [9] and integration of the predictive model in a smart solar monitoring software, including an intervention management system integrated with alarm handling and a business intelligence based reporting tool from intervention up to portfolio level [10].

## 5 ACKNOWLEDGMENTS

Authors thank Prof. Mauro Tucci and Prof. Emanuele Crisostomi from Destec Department of University of Pisa for fruitful discussions and helpful suggestions.